\documentclass[11pt]{article}

\usepackage{acl}

\usepackage{times}
\usepackage{latexsym}
\usepackage[T1]{fontenc}
\usepackage[utf8]{inputenc}
\usepackage{microtype}
\usepackage{inconsolata}
\usepackage{graphicx}
\usepackage{booktabs}
\usepackage{amsmath}
\usepackage{amssymb}
\usepackage{multirow}
\usepackage{placeins}

\graphicspath{{./}{figures/}}

\title{Temporal Taxation Compounds Under Post-Training Compression of Whisper Models}
 
\author{
  Srishti Ginjala\textsuperscript{1} \qquad
  Eric Fosler-Lussier\textsuperscript{1} \\[0.2em]
  Christopher W. Myers\textsuperscript{2} \qquad
  Srinivasan Parthasarathy\textsuperscript{1} \\[0.3em]
  \textsuperscript{1}The Ohio State University, Columbus, OH, USA \qquad
  \textsuperscript{2}Air Force Research Laboratory, USA \\[0.2em]
  \texttt{ginjala.1@osu.edu} \quad
  \texttt{\{fosler,srini\}@cse.ohio-state.edu} \\
  \texttt{christopher.myers.29@us.af.mil}
}

\begin{document}
\raggedbottom
\maketitle

\begin{abstract}
Automatic speech recognition models are audited for demographic fairness at full precision, yet the models that ship to production have been quantized, pruned, and distilled. We ask whether \emph{post-training weight compression}, which alters model weights rather than the audio signal or its feature representation, redistributes error burden across demographic groups. Across the Whisper family on Fair-Speech, Common Voice 25, and AfriSpeech-200, 50\% Wanda pruning of Whisper-large-v3 sharply widens the Black/AA-vs-Asian temporal-taxation differential on Fair-Speech: the absolute word-error-rate gap between the worst- and best-served groups more than doubles; at an assumed cost of five seconds of correction effort per transcription error this is a rise from 30 to 64 seconds of correction time per minute of speech. This $+111\%$ relative increase is invariant to the assumed per-error cost, survives an audio-quality control, and is only partly mitigated by beam-search decoding, which still leaves an $+86\%$ increase. At edge model size, INT4 HQQ quantization compounds catastrophic transcript loops on West African accents by factors of five to seven. Distillation, by contrast, narrows demographic gaps in 21 of 27 evaluated settings (teacher-student pair, precision, and dataset), with the exceptions concentrated on a single model pair. We cast the temporal-taxation construct of \citet{choi2025temporal} as a quantitative metric, and show that single-snapshot fairness audits on full-precision models do not capture the deployment-time burden that compression places on already-marginalized speakers.
\end{abstract}

\section{Introduction}
\label{sec:intro}

Regulatory fairness audits of automatic speech recognition (ASR) systems typically evaluate a full-precision model and report word-error-rate gaps across demographic groups \citep{koenecke2020racial,tatman2017gender,veliche2024fairspeech}. However, the system deployed in production, especially on edge devices, is rarely the same as the one audited. These systems are often reduced-order distilled or quantized models. For instance, the Whisper-large-v3 model alone is downloaded over four million times each month from the Hugging Face Hub and has numerous derivative checkpoints, many of which are quantized to INT8 or INT4, magnitude-pruned, or distilled. This paper examines how the per-group failure profile changes when such \emph{post-training weight compression} is applied, by which we mean quantization, pruning, and distillation of model weights, as distinct from compression of the audio signal or of its feature representation.

The question is not academic. In transcription, captioning, and dictation workflows, recognition errors are typically surfaced to human reviewers for correction, so a demographic gap in word error rate becomes a gap in correction labor: worse-served speakers spend more time repairing transcripts for the same minute of speech. The relative compounding we report does not depend on this assumption (Section~\ref{sec:methods:metrics}). \citet{choi2025temporal} formalize this burden as \emph{temporal taxation}, which we adopt as our headline metric because it expresses ASR errors as a concrete time cost to the speaker rather than an abstract percentage. Compression theory in vision \citep{hooker2019compressed,hooker2020characterising} and multilingual NLP \citep{ogueji2022compression,ahia2021lowresource} predicts that pruning and quantization disproportionately damage the long tail of input distributions; text-LLM work reports that quantization effects on bias are complex and unpredictable across techniques \citep{kirsten2025inference}. Whether ASR demographic groups behave as this long tail under compression is therefore an empirical question with no confidently predicted direction. Our findings diverge from the naive long-tail extrapolation in three ways. Pruning compounding is monotonic in \emph{increasing} model capacity, from $-0.56$ at \textit{whisper-tiny} to $+0.58$ at \textit{whisper-large-v3} (Section~\ref{sec:discussion}), the opposite of the pattern reported in vision. Distillation narrows demographic gaps in 21 of 27 settings tested (Section~\ref{sec:distillation}), against the prediction that compression damages the tail. And NF4 and HQQ at the same bit-width redistribute error to different groups (Section~\ref{sec:quantization:wer}), so the extrapolation does not select a direction even within one bit-width.

Prior ASR fairness work treats the model as fixed: per-model audits at FP32 or FP16, no compression \citep{koenecke2020racial,tatman2017gender,harris2024modeling,veliche2024fairspeech,ezema2025cascading,cunningham2024impacts,koenecke2024careless}. \citet{cheng2026diversity} introduces a semantic-axis audit but evaluates only at FP32. Prior ASR compression work has rarely examined fairness. The main exception is \citet{ferraz2024multilingual}, which measures how 8-bit quantization affects multilingual demographic gaps in Whisper, but only at 8-bit and without English dialect or accent axes. The closest LLM-side work is \citet{kirsten2025inference}, who find that inference-acceleration effects on bias in text models are mixed and method-specific.

Our contributions are fourfold. (i)~We present the first systematic study of post-training pruning and ASR fairness: 50\% Wanda pruning \citep{sun2024wanda} of Whisper-large-v3 more than doubles the Black/AA-vs-Asian temporal-taxation differential on Fair-Speech ($+111\%$), an effect invariant to the assumed per-error cost and robust to an audio-quality control (Section~\ref{sec:pruning}), and reduced but not removed by beam-search decoding, which still leaves a $+86\%$ increase (Section~\ref{sec:beam}). (ii)~We extend the 8-bit fairness analysis of \citet{ferraz2024multilingual} to sub-8-bit precision on English dialect and accent: at edge model size, INT4 HQQ quantization multiplies catastrophic-loop rates on West African accents by factors of five to seven, and WER compounding is capacity-dependent and recipe-specific (Section~\ref{sec:quantization}). (iii)~We operationalize the temporal-taxation construct of \citet{choi2025temporal}, originally a conceptual proposal, as a reporting convention for ASR compression audits. The metric is a linear rescaling of per-group WER and we do not claim the arithmetic as a contribution; what it buys is a deployment-interpretable time unit and, because of the linearity, a relative compounding measure that is exactly invariant to the cost-per-error assumption (Section~\ref{sec:methods:metrics}). (iv)~We report a counter-expectation result: distillation does not compound demographic gaps but narrows them in most settings tested, with one structured exception (Section~\ref{sec:distillation}).

Our results are within the Whisper family, chosen to isolate the effect of compression from confounds with architecture, tokenizer, and training data (Section~\ref{sec:methods:models}); a probe on IBM Granite-4.0-1b-speech (Appendix~\ref{app:granite}) tests one direction of cross-family generalization.

\section{Related work}
\label{sec:related}

\paragraph{ASR Fairness Audits.} A line of work documents demographic disparities in ASR at fixed precision. \citet{koenecke2020racial} reports a two-fold racial gap across five commercial systems, and \citet{tatman2017gender} surfaces a gender-race interaction in early YouTube and Bing captions. \citet{martin2020habitual} attribute part of the racial gap to mishandling of habitual \textit{be} and other African American English features, and \citet{harris2024modeling} extend the gender-dialect interaction to modern ASR. \citet{ezema2025cascading} catalog cascading effects of ASR bias in spoken-language interfaces, and \citet{cunningham2024impacts} document downstream impacts on speakers of African American Language. \citet{veliche2024fairspeech} release the Fair-Speech benchmark. \citet{koenecke2024careless} catalog Whisper hallucination harms that motivate the loop-rate axis tracked here. \citet{cheng2026diversity} most recently introduced a semantic-axis fairness metric. All of this work treats the model as fixed: it audits \emph{which} model fails on whom, not \emph{what happens to that failure profile after compression}.

\paragraph{Compression and Bias.} Some researchers observe that compression damages the long tail of input distributions: \citet{hooker2019compressed,hooker2020characterising} demonstrate this for vision models, and \citet{ogueji2022compression,ahia2021lowresource} show the same for multilingual NLP. \citet{kirsten2025inference} find that inference-acceleration effects on LLM bias are "complex and unpredictable" across techniques. On the ASR dimension, \citet{ferraz2024multilingual} reports that 8-bit quantization of Whisper amplifies model-related biases (resourcefulness, model size) on FLEURS and Common Voice 13 while leaving speaker-related biases (gender, age) roughly stable. Still, the analysis is limited to 8-bit precision and does not examine English dialect or accent. \citet{feng2025edgeasr} benchmark low-bit Whisper for edge deployment and \citet{andreyev2025whisperquant} compare quantization recipes; neither studies demographic fairness. The vision and text literatures predict compounding; the speech literature has examined only 8-bit quantization, and only on multilingual axes.

\paragraph{Adjacent work on Whisper deployment-time fairness.} \citet{ginjala2026prior} study how acoustic input degradation and decoder architecture interact with ASR demographic fairness. The present work is orthogonal: it perturbs model weights, through quantization, pruning, and distillation, rather than the acoustic input, and the two experimental matrices share no common (model, condition) cell.

\section{Methods}
\label{sec:methods}

\subsection{Models}
\label{sec:methods:models}

The main analysis covers eight Whisper-family models: five Whisper backbones (\textit{whisper-tiny}, \textit{whisper-base}, \textit{whisper-small}, \textit{whisper-medium}, \textit{whisper-large-v3}; 39M to 1.55B parameters) and three Distil-Whisper students paired with their teachers (\textit{distil-small.en}, \textit{distil-medium.en}, \textit{distil-large-v3}) \citep{radford2023whisper,gandhi2023distilwhisper}. Although whisper-large-v3 no longer leads the Open ASR Leaderboard, the Whisper family spans the widest publicly available capacity range and has the most calibrated demographic-fairness baselines in prior work \citep{koenecke2020racial,ferraz2024multilingual}. We exclude Whisper-large-v3-Turbo because Turbo is itself a layer-pruned compression of large-v3 and would confound the distillation arm.

Restricting the main analysis to one family is deliberate: a cross-family comparison would confound the compression effect with architecture, tokenizer, training data, and audio encoder, the confound that limits prior multi-family ASR fairness audits. The Granite-4.0 probe in Appendix~\ref{app:granite} gives a one-architecture check on the FP16/INT8 gap structure, but does not test the pruning or INT4 claims cross-family.

\subsection{Compression methods}
\label{sec:methods:compression}

\paragraph{Quantization.} The sweep covers four precisions per model: FP16 as the deployment-realistic reference, INT8 \citep{dettmers2022llmint8}, INT4 NF4 (NormalFloat4), and INT4 HQQ (Half-Quadratic Quantization; \citealp{badri2023hqq}). NF4 and HQQ are both data-free; including both isolates the optimization-recipe axis while holding bit-width and calibration-data status constant. Adapting HQQ to Whisper's encoder-decoder required an implementation fix, which we verified on a held-out set using FP16. GPTQ \citep{frantar2023gptq} and AWQ \citep{lin2023awq} are excluded on implementation-availability grounds rather than algorithmic ones: both algorithms are architecture-agnostic, but the reference libraries we tested (\texttt{gptqmodel}, \texttt{autoawq}) expose per-architecture calibration classes that do not include encoder-decoder speech models. Appendix~\ref{app:excluded} documents the specific blockers and notes that torchao provides a module-level configuration path we did not pursue.

\paragraph{Pruning.} We apply Wanda 50\% unstructured pruning \citep{sun2024wanda}, extended to cover the linear layers of Whisper's encoder, decoder, and decoder cross-attention. Calibration uses 128 utterances from LibriSpeech train-clean-100, held out from every test set used for fairness claims. SparseGPT \citep{frantar2023sparsegpt} requires per-architecture model classes that do not currently exist for encoder-decoder speech models.

\paragraph{Distillation.} Distil-Whisper checkpoints are paired with the corresponding Whisper teacher \citep{gandhi2023distilwhisper}, evaluated at each precision in the quantization sweep.

\subsection{Datasets and demographic axes}
\label{sec:methods:datasets}

Three benchmarks carry the fairness claims, with LibriSpeech test-clean held in reserve as an aggregate-WER reference only. Fair-Speech \citep{veliche2024fairspeech} provides controlled read voice-assistant prompts (26{,}471 utterances, 593 speakers, demographic axes for ethnicity, age, gender, geography, and native-English status). Common Voice 25 supplies the English test split (16{,}398 clips) with accent labels populated for seven groups at $n \geq 50$ utterances (us 1343, indian 527, england 437, canada 116, australia 100, scotland 53, african 53). AfriSpeech-200 provides 6{,}318 test utterances over 44 fine-grained African accents with $n \geq 50$.

\subsection{Metrics}
\label{sec:methods:metrics}

Per-utterance metrics include word error rate (WER), deletion rate, insertion rate, and a per-utterance loop flag set when a hypothesis contains more than five 5-gram repetitions or when the hypothesis-to-reference length ratio exceeds three (thresholds set before the full sweep ran; see Appendix D). For each (model, configuration, dataset) cell, we aggregate WER by demographic group and report the utterance-weighted mean WER with $B = 200$ bootstrap 95\% confidence intervals. Per-cell fairness is summarized by the max-min ratio $\mathrm{MMR} = \max_g \mathrm{WER}_g / \min_g \mathrm{WER}_g$ and the compounding index $(\mathrm{MMR}_{\mathrm{compressed}} -\allowbreak \mathrm{MMR}_{\mathrm{FP16}})\allowbreak/\allowbreak\mathrm{MMR}_{\mathrm{FP16}}$.

The headline metric is the temporal-taxation differential adapted from \citet{choi2025temporal}. For demographic group $g$ in a cell with audio of mean rate $\mathit{wpm}$ words per minute,
\[
T_g \;=\; \mathrm{WER}_g \cdot C \cdot \mathit{wpm},
\]
where $C$ is the assumed cost per error in seconds. $T_g$ is a linear rescaling of $\mathrm{WER}_g$; we do not present the transformation itself as a contribution. Its role is to express the gap in a deployment-interpretable unit and to separate the cost-dependent part of the claim from the cost-independent part. We report the worst-versus-best group difference and the relative change in this difference under compression. Because $T_g$ is linear in $C$, the relative change in the differential is exactly invariant in $C$; only absolute anchors depend on $C$. We anchor at $C = 5$\,s/edit and report sensitivity over $C \in \{2, 5, 8\}$\,s/edit (Appendix B). The linear-cost assumption applies to per-token errors; loop-induced cost, which scales non-linearly with hypothesis length, is tracked separately on the loop-rate axis (Section~\ref{sec:quantization:loops}).

\subsection{Statistical methodology}
\label{sec:methods:stats}

We test paired per-utterance WER differences with a paired permutation test ($B = 2{,}000$ permutations per cell, two-sided) and adjust within each cell with the Benjamini-Hochberg procedure across demographic groups \citep{benjamini1995controlling}. Confidence intervals for per-group means use $B = 200$ percentile bootstrap samples. Ranking stability of models by MMR across precisions is summarized by Kendall's $\tau$. Following \citet{li2024provenance}, we test for an audio-quality confound on Fair-Speech by computing a per-utterance SNR proxy, fitting an ordinary-least-squares regression of WER on SNR, and re-testing the demographic F-statistic after the SNR adjustment (full output in Appendix~C).

\section{Pruning effects on demographic temporal taxation}
\label{sec:pruning}

\begin{figure*}[!tbp]
  \centering
  \includegraphics[width=0.88\textwidth]{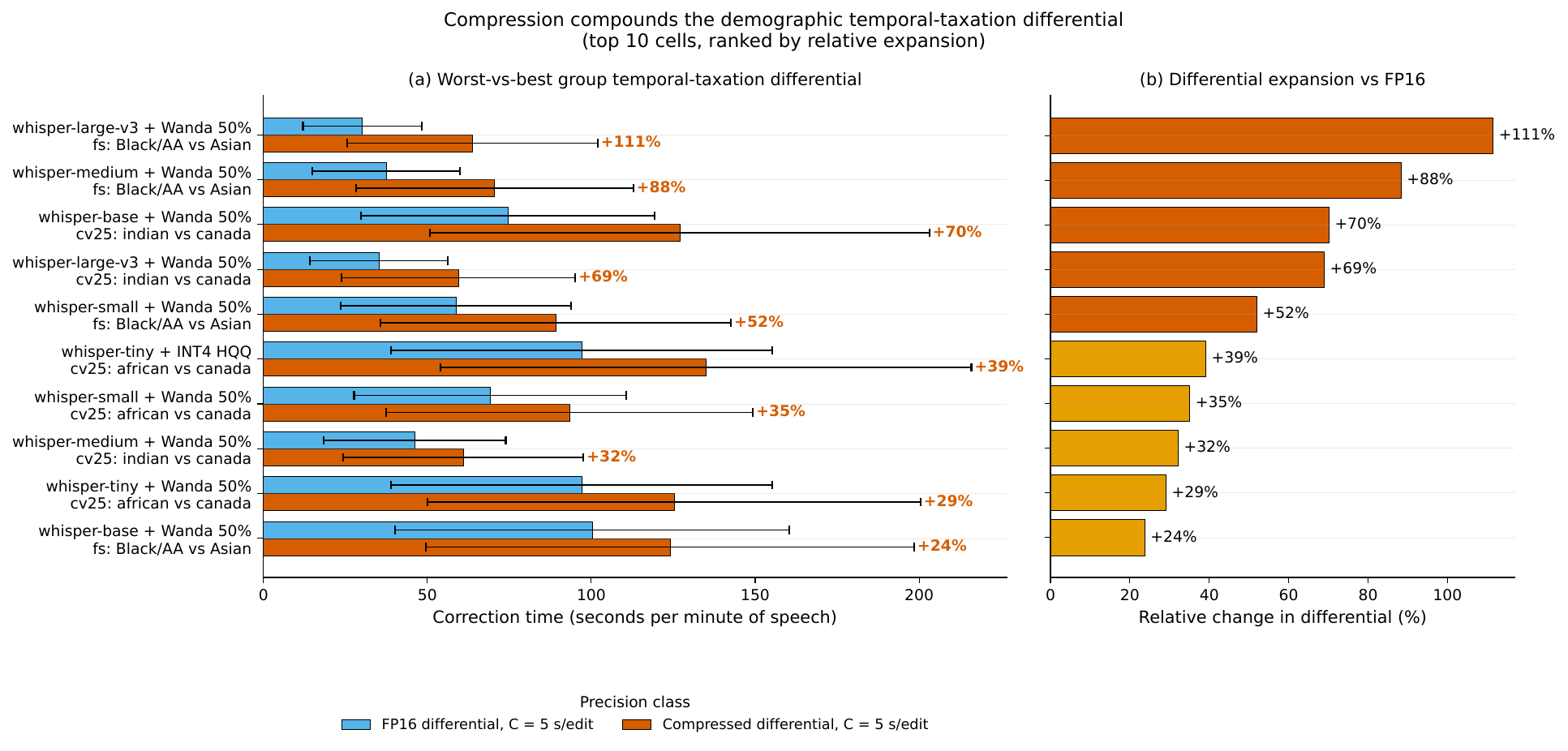}
  \caption{Temporal-taxation differential under pruning and quantization, between the worst- and best-served demographic groups, top ten cells by relative compounding. \textbf{(a)}~Absolute differential in seconds of correction per minute of speech, FP16 (blue) versus compressed (vermilion), at five seconds per error; horizontal extensions show the differential range when the per-error cost varies from two to eight seconds. \textbf{(b)}~Relative change in the differential under compression, which is invariant to the per-error cost assumption. The headline cell is \textit{whisper-large-v3} with 50\% Wanda pruning on Fair-Speech (Black/AA vs Asian, $+111\%$).}
  \label{fig:taxation}
\end{figure*}

\subsection{Headline result}
\label{sec:pruning:headline}

Deploying \textit{whisper-large-v3} with 50\% Wanda unstructured pruning on Fair-Speech raises the Black/AA-vs-Asian temporal-taxation differential from 30.14 to 63.73 seconds of correction time per minute of speech at $C = 5$\,s/edit, a relative increase of 111.4\% (Figure~\ref{fig:taxation}). The differential more than doubles, and the relative change is exactly invariant in $C$ (sensitivity table in Appendix~B), so the $+111\%$ claim does not depend on the cost-per-error assumption. The asymmetry is concentrated in the worse-served group: under Wanda, Black/AA WER rises by $+7.89$ percentage points (BH-adjusted $p \approx 0$, $n = 7782$), while Asian WER rises by only $+0.43$ points ($p_{\mathrm{BH}} = 0.006$, $n = 3853$).

\begin{figure*}[!tbp]
  \centering
  \includegraphics[width=\textwidth]{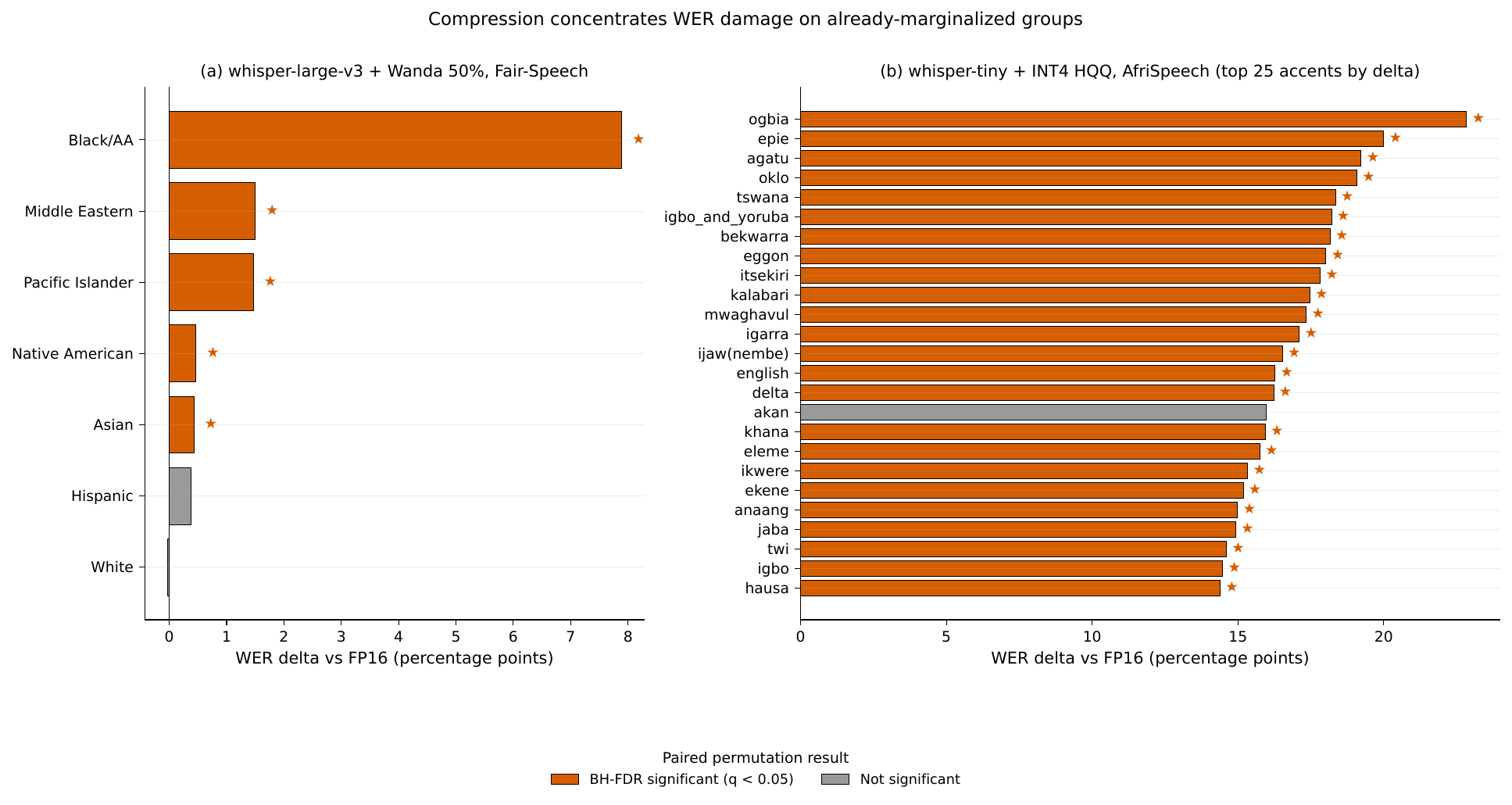}
  \caption{Per-group WER damage under pruning (a) and quantization (b) concentrates on already-marginalized groups. Per-group WER delta (compressed minus FP16, percentage points) with paired-permutation BH-FDR-significant cells in vermilion (q$<$0.05). (a)~Fair-Speech ethnicity under \textit{whisper-large-v3}~+~50\% Wanda: Black/AA $+7.89$\,pp; Hispanic and White below the BH threshold. (b)~AfriSpeech top-25 accents by delta under \textit{whisper-tiny}~+~INT4 HQQ; all but akan BH-significant.}
  \label{fig:forest}
\end{figure*}

The pattern replicates across cells: \textit{whisper-large-v3} + Wanda widens the indian-vs-canada Common Voice 25 differential by $+68.9\%$, and \textit{whisper-small} + Wanda widens Black/AA-vs-Asian on Fair-Speech by $+51.9\%$ and African-vs-Canadian on Common Voice 25 by $+35.1\%$ (absolute s/min anchors for all cells in Appendix~\ref{app:taxation-sensitivity}). Nine of the top ten cells in Figure~\ref{fig:taxation} are Wanda; pruning produces the largest demographic-redistribution effects in the experimental matrix. Figure~\ref{fig:forest} panel (a) shows the per-group breakdown: under \textit{whisper-large-v3} + Wanda, Black/AA stands alone at $+7.89$\,pp while Asian, Native American, Pacific Islander, and Middle Eastern speakers cluster below $+1.5$\,pp.

\subsection{Audio-quality confound}
\label{sec:pruning:snr}

A standing concern in ASR fairness measurement is that demographic groups are confounded with recording quality, so observed WER gaps reflect microphone or environmental differences rather than accent or dialect \citep{li2024provenance}. We compute a per-utterance SNR proxy on Fair-Speech, regress per-utterance WER on SNR, and re-test the demographic F-statistic on the residual. The ethnicity effect survives the SNR control in 34 of 34 (model, precision) Fair-Speech cells. After SNR adjustment, the demographic MMR is larger than the raw MMR in every cell, by 15.0\% on average (median 13.9\%); the SNR control widens the gap rather than narrowing it. The pruning compounding result is therefore not explained by SNR-measurable audio-quality variation; the raw corpus WER mildly understates the disparity rather than overstating it. SNR is a partial proxy: reverberation and non-linear distortion are not captured, and residual confounding from those sources cannot be excluded (Limitations).

\section{Quantization effects across capacity and method}
\label{sec:quantization}

\subsection{WER compounding and capacity scaling}
\label{sec:quantization:wer}

\begin{figure*}[!tbp]
  \centering
  \includegraphics[width=0.88\textwidth]{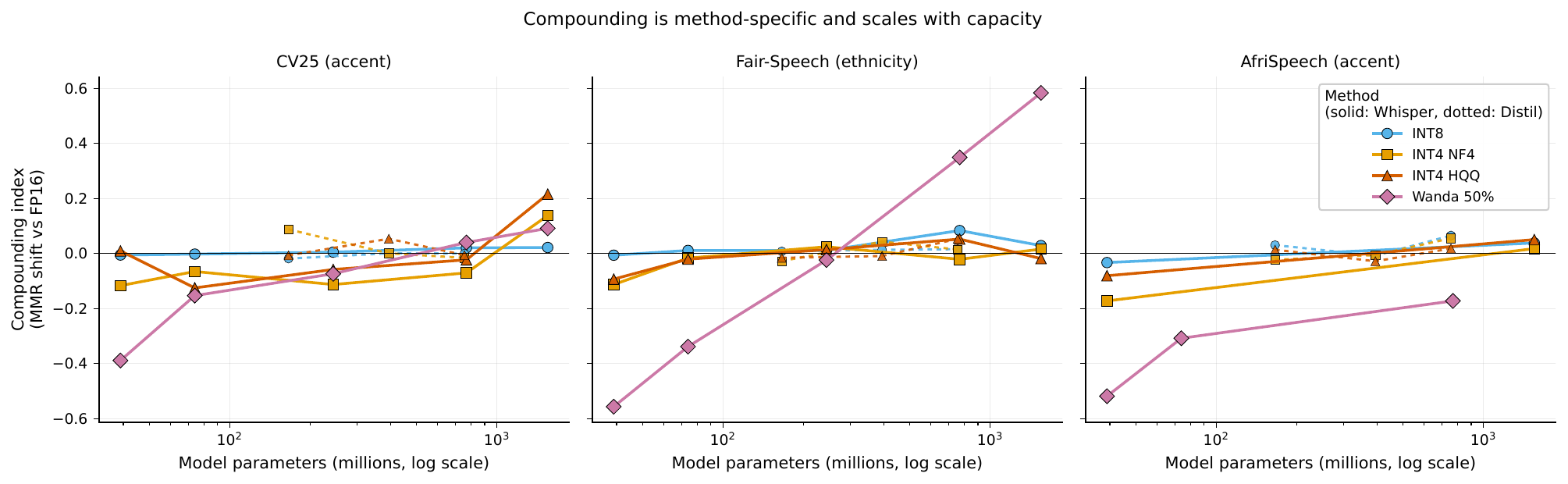}
  \caption{Quantization and pruning compounding is method-specific and scales with capacity. Compounding index (MMR shift relative to FP16) versus Whisper parameter count (log scale); one panel per dataset; each panel covers four methods (INT8, INT4 NF4, INT4 HQQ, and 50\% Wanda pruning). Solid: Whisper backbones; dotted: Distil-Whisper. The Fair-Speech Wanda curve (purple) is monotonic from $-0.56$ at \textit{whisper-tiny} to $+0.58$ at \textit{whisper-large-v3}, opposite in direction to sub-8-bit quantization. AfriSpeech Wanda is evaluated only at \textit{whisper-tiny}, \textit{whisper-base}, and \textit{whisper-medium}.}
  \label{fig:scaling}
\end{figure*}

WER compounding under quantization is capacity-bounded (Figure~\ref{fig:scaling}). At \textit{whisper-tiny} on Common Voice 25, INT4 HQQ produces 7 of 7 BH-significant per-accent deltas, all positive, ranging from $+5.45$\,pp (Canadian) to $+13.19$\,pp (African). At \textit{whisper-large-v3} on Common Voice 25, INT4 HQQ produces a smaller MMR shift ($+0.216$), and only the British English accent reaches BH significance ($+0.73$\,pp); the MMR change at this size is driven by which accent occupies the extremes of the per-group WER distribution rather than by large per-group magnitudes. Four mid-size Common Voice 25 quantization cells show the opposite pattern, with compounding index between $-0.13$ and $-0.06$: MMR narrows because the best-served accent degrades proportionally harder than the worst-served one. Under \textit{whisper-base} + INT4 HQQ, Canadian WER rises 28\% (9.6\% to 12.3\%) while African WER rises 12\% (24.9\% to 27.8\%).

NF4 and HQQ at the same bit-width produce distinguishable redistributions. At \textit{whisper-tiny} both recipes compound on the WER axis, but unevenly: on Common Voice 25, HQQ yields 7 of 7 BH-significant per-accent deltas against 5 of 7 for NF4, and on the loop-rate axis HQQ reaches significance on 7 of 62 AfriSpeech accents and 1 of 7 Fair-Speech groups where NF4 reaches 1 of 62 and 0 of 7. NF4 nonetheless flattens MMR more at this size (Common Voice 25 compounding index $-0.117$ for NF4 against $+0.010$ for HQQ), because it damages the best-served accent proportionally harder. At \textit{whisper-large-v3} the per-group picture inverts on AfriSpeech, where NF4 reaches BH significance on 3 of 63 accents and HQQ on none, while on Common Voice 25 HQQ shifts MMR further than NF4 ($+0.216$ against $+0.138$). The choice of 4-bit recipe therefore carries a fairness consequence as well as an accuracy consequence, consistent with \citet{kirsten2025inference}'s finding that quantization method interacts with bias on text LLMs. Figure~\ref{fig:forest} panel (b) shows the per-accent evidence for the strongest WER-compounding cell.

\subsection{Loop-rate compounding}
\label{sec:quantization:loops}

Aggregate WER is a poor summary of user-experienced cost when a fraction of utterances produce catastrophic transcripts: long hallucination loops, sustained 5-gram repetitions, and length-blowups that force minutes of correction effort on a single sentence \citep{koenecke2024careless}. We track the loop axis separately. A hypothesis is flagged as a loop when it contains more than five 5-gram repetitions or its length exceeds three times the reference length; the thresholds are set in advance of the full sweep (Appendix~D).

\begin{figure*}[!tbp]
  \centering
  \includegraphics[width=\textwidth]{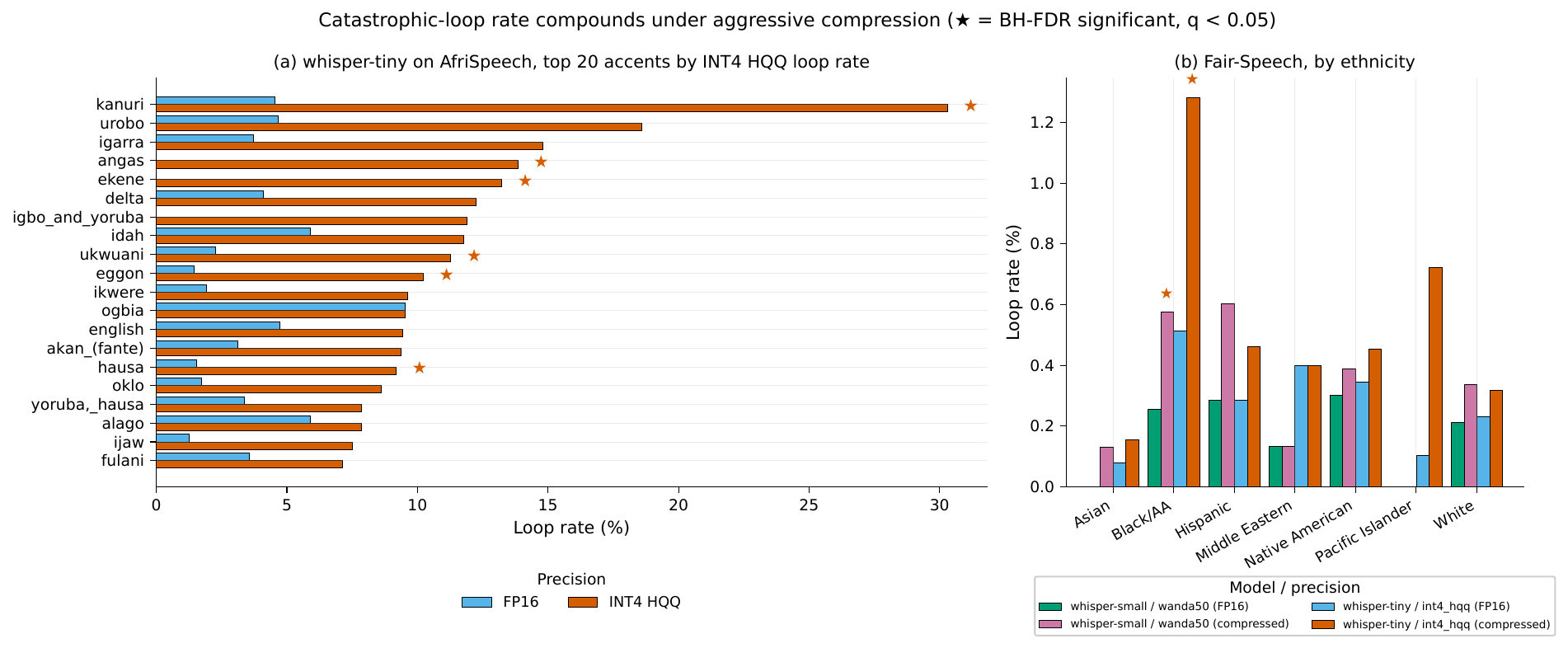}
  \caption{Catastrophic-loop rate under quantization and pruning compounds at edge capacity ($\bigstar$~=~BH-FDR significant, $q < 0.05$). (a)~\textit{whisper-tiny} on AfriSpeech, top-20 accents by INT4 HQQ loop rate. (b)~Fair-Speech ethnicity loop rates for \textit{whisper-tiny}~+~INT4 HQQ and \textit{whisper-small}~+~50\% Wanda.}
  \label{fig:loops}
\end{figure*}

Loop rate compounds (increases) sharply with edge model size for West African accents. Under \textit{whisper-tiny} with INT4 HQQ on AfriSpeech, seven of 62 accents become BH-significantly worse than FP16 in paired permutation (Figure~\ref{fig:loops}). Kanuri (n=66) moves from 4.55\% to 30.30\% loop rate, a 6.67-fold increase ($p_{\mathrm{BH}} = 2.68 \times 10^{-3}$). Hausa (n=196) moves from 1.53\% to 9.18\%, a 6.0-fold increase ($p_{\mathrm{BH}} = 1.43 \times 10^{-2}$). Yoruba (n=648) moves from 1.23\% to 6.33\%, a 5.12-fold increase ($p_{\mathrm{BH}} = 8.62 \times 10^{-5}$). The Fair-Speech replication is in the same direction: under \textit{whisper-tiny} + INT4 HQQ, Black/AA loop rate rises from 0.51\% to 1.28\% (a 2.5-fold increase, $n=7807$, $p_{\mathrm{BH}} = 2.46 \times 10^{-6}$). White and other ethnic groups on Fair-Speech show smaller, non-significant changes.

Since the \emph{relative} compounding metric controls for the FP16 baseline within each accent, these results isolate the quantization-induced increase in loop rate from baseline accent difficulty. A Kanuri 4.55\% loop rate at FP16 is already high; the claim is that INT4 HQQ multiplies that rate by 6.67$\times$, not that the absolute rate is high. The signal weakens at larger model sizes: at \textit{whisper-large-v3} + INT4 HQQ on AfriSpeech, no accent reaches BH significance for loop-rate compounding (Appendix~A). The loop-rate finding therefore applies to \emph{edge}-deployable models, where INT4 is the realistic deployment precision.

\subsection{Ranking stability}
\label{sec:quantization:ranking}

Model rankings by MMR are highly stable across precisions. Kendall's $\tau$ between the FP16 ranking and the compressed-precision ranking is 1.000 in 11 of 12 (dataset, precision) cells, with one adjacent-pair swap at Fair-Speech + INT4 HQQ ($\tau = 0.929$); full table in Appendix~\ref{app:kendall}. A practitioner choosing a model on a full-precision audit can rely on the relative ordering surviving deployment at sub-8-bit precision; the absolute fairness gap, in contrast, does not survive (Sections~\ref{sec:pruning} and~\ref{sec:quantization:wer}).

\subsection{Beam-search decoding does not mitigate compounding}
\label{sec:beam}

A natural mitigation to consider is whether beam-search decoding compensates for compression error. We rerun two models (\textit{whisper-small}, \textit{whisper-large-v3}) at two precisions (FP16, INT4 NF4) on Fair-Speech and Common Voice 25 with a beam width of 5, producing 8 quantization ablation cells against the greedy baseline, and separately rerun the headline pruning cell (\textit{whisper-large-v3} + 50\% Wanda on Fair-Speech) at the same beam width. In six of the eight cells the beam-5 MMR moves by less than $0.2$ from its greedy counterpart, and in three of the four (model, dataset) pairs beam-5 leaves the NF4 compounding index unchanged or slightly reduces it. In the fourth pair, \textit{whisper-large-v3} + INT4 NF4 on Common Voice 25, beam-5 \emph{amplifies} the compounding index relative to greedy ($+0.296$ vs $+0.138$). Loop rate increases under beam-5 in eight of eight cells, with the largest absolute jump on \textit{whisper-large-v3}. Beam search is not a deployment-time fairness fix for quantization compounding; it can amplify it.

On the pruning arm, beam-5 helps but does not resolve the disparity. Under \textit{whisper-large-v3} + 50\% Wanda on Fair-Speech, the compounding index falls from $+0.584$ under greedy decoding to $+0.468$ under beam-5, and the Black/AA-vs-Asian temporal-taxation differential rises from 27.95 to 51.91\,s/min rather than from 30.14 to 63.73\,s/min, so the headline increase moves from $+111\%$ to $+86\%$. The redistribution is unchanged in shape: Black/AA still carries by far the largest degradation ($+5.63$\,pp under beam-5 against $+7.89$\,pp under greedy), the next-largest group delta is under $+1.3$\,pp, and six of seven ethnicity groups reach BH significance under beam-5 against five of seven under greedy, because beam-5's lower variance makes the smaller deltas detectable. Beam search buys back roughly a fifth of the pruning compounding at roughly twice the inference cost; it does not make pruning demographically neutral.

\section{Distillation effects on demographic gaps}
\label{sec:distillation}

\begin{figure*}[!tbp]
  \centering
  \includegraphics[width=0.88\textwidth]{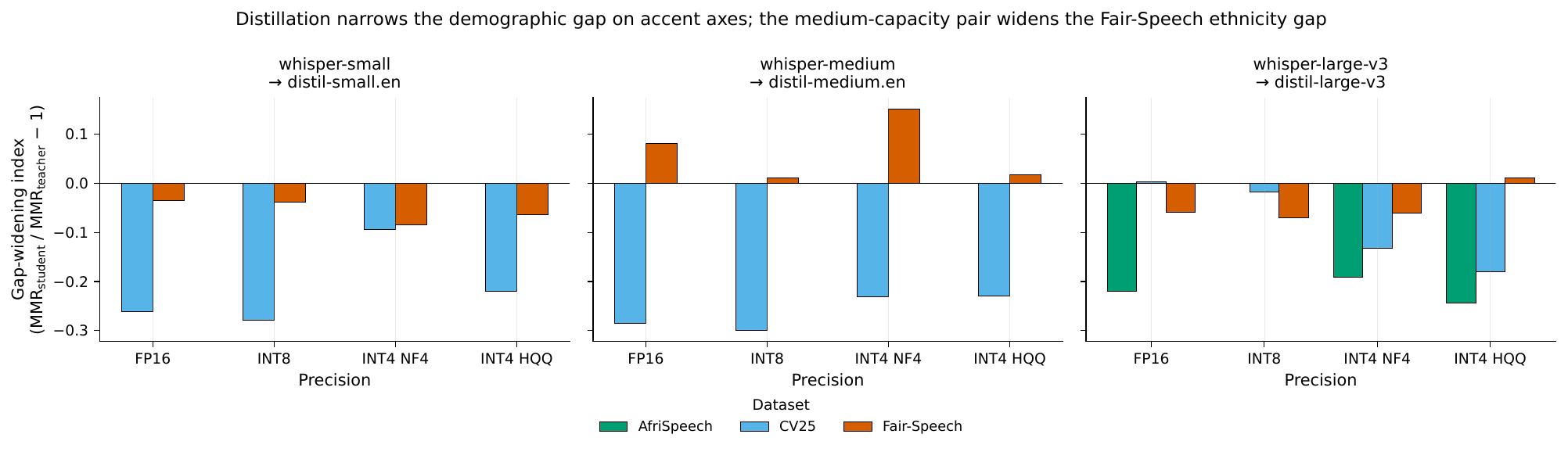}
  \caption{Distillation narrows the demographic gap on accent axes but widens the Fair-Speech ethnicity gap for the whisper-medium $\rightarrow$ distil-medium.en pair (center panel). Gap-widening index $= (\mathrm{MMR}_{\mathrm{student}} / \mathrm{MMR}_{\mathrm{teacher}}) - 1$, by precision; one panel per teacher-student pair; bars colored by dataset; negative values indicate narrowing.}
  \label{fig:rq3}
\end{figure*}

At the deployment-relevant model size, distillation narrows the demographic gap on the accent axis: six of the seven \textit{distil-large-v3} cells on AfriSpeech and Common Voice 25 reduce MMR relative to their \textit{whisper-large-v3} teacher (the exception, Common Voice 25 at FP16, is a $+0.003$ shift within noise), and on AfriSpeech under INT4 HQQ (the only pair and precision evaluated at this scale) 57 of 63 accents narrow with BH significance. The direction replicates across the matrix: over 27 (teacher, student, precision, dataset) cells, distillation narrows demographic fairness gaps in 21 cells, shows systematic widening on 4, and shows near-zero shifts within noise on the remaining 2 (Figure~\ref{fig:rq3}). The widening is structured rather than scattered, concentrated on the \textit{whisper-medium} $\rightarrow$ \textit{distil-medium.en} pair on Fair-Speech ethnicity at every precision (gap-widening index $+0.011$ to $+0.151$); the two near-zero shifts (distil-large-v3 + Fair-Speech + INT4 HQQ at $+0.010$, distil-large-v3 + Common Voice 25 + FP16 at $+0.003$) are within measurement noise.

Aggregate WER rises uniformly under distillation across all 27 cells, with relative student-vs-teacher WER deltas ranging from $+0.07\%$ on \textit{distil-small.en} + INT4 HQQ + Common Voice 25 to $+31.4\%$ on \textit{distil-medium.en} + INT4 HQQ + Fair-Speech. The narrowing is a relative-parity result rather than an absolute improvement: distillation raises every group's WER, but the worst-served group's WER scales sub-linearly with the aggregate increase, so the ratio compresses. Distillation is therefore fairness-preserving on the accent axis, with one exception (the medium-capacity teacher-student pair on read-speech ethnicity). Unlike pruning and sub-8-bit quantization, distillation does not concentrate cost on already-marginalized groups, even though it makes every group worse in absolute terms.

\paragraph{Candidate mechanisms.} Why distillation narrows where pruning compounds is open. Training on teacher pseudo-labels averages the student's loss over the teacher's full input distribution, which may regularize against the worst-group behavior the teacher already exhibits, and may also push the student toward a smoother decision boundary that is less peaked on majority-group features. Adjudicating between these would require training-data attribution or controlled pseudo-label experiments, which we leave to future work.

\section{Concluding Remarks and Discussion}
\label{sec:discussion}

Across the three compression families, the deployment-vs-benchmark gap is asymmetric. Pruning compounds the demographic temporal-taxation differential at the deployment scale; distillation narrows it on the accent axis (with one structured-ethnicity exception); and quantization sits between, capacity-bounded and recipe-specific. The three families are therefore not interchangeable from a fairness perspective. A practitioner choosing between 50\% Wanda pruning and INT4 HQQ quantization to fit Whisper-large-v3 on a mobile device is making a fairness decision: Wanda more than doubles the Black/AA-vs-Asian Fair-Speech differential, while sub-8-bit quantization at the same model size shifts MMR by less than 0.25 and reaches BH significance on only one accent group (England, $+0.73$\,pp). This characterization applies to the data-free 4-bit recipes and to Wanda; GPTQ, AWQ, and SparseGPT, which dominate practitioner deployment in many settings, fall outside this matrix on implementation-availability grounds: the algorithms are architecture-agnostic, but the reference libraries expose per-architecture calibration classes that do not include encoder-decoder speech models (Appendix~\ref{app:excluded}).

Sub-8-bit quantization compounding on the WER axis fades as model size grows: at \textit{whisper-large-v3}, INT4 HQQ yields a Common Voice 25 MMR shift of $+0.216$ and smaller Fair-Speech shifts concentrated on specific groups. The Wanda axis moves in the opposite direction: across the five Whisper backbones, the Fair-Speech compounding index is strictly monotonic in parameter count, sweeping from $-0.56$ at \textit{whisper-tiny} through $-0.34, -0.02, +0.35$ to $+0.58$ at \textit{whisper-large-v3} (Figure~\ref{fig:scaling}), where the $+111\%$ taxation differential is the endpoint of the curve. At edge capacity, Wanda flattens MMR by crushing the best-served group harder than the worst-served; compounding emerges at medium capacity. One candidate mechanism, consistent with prior compression work in vision \citep{hooker2019compressed,hooker2020characterising} and multilingual NLP \citep{ogueji2022compression,ahia2021lowresource}, is that Wanda's magnitude $\times$ activation pruning criterion preferentially removes weights serving low-frequency input distributions, which in Whisper training data correspond to the demographic groups already worst-served at full precision. Quantization, by contrast, scales magnitudes uniformly within each quantization bucket and does not selectively prune the long tail. Our experiments establish the empirical asymmetry but cannot adjudicate the mechanism: the deletion-rate decompositions in Appendix~\ref{app:full-results} are consistent with a long-tail account but are not decisive. A decisive test would need weight-level attribution to training-data subpopulations, or a controlled experiment varying long-tail representation in the calibration set.

Because the temporal-taxation differential is linear in $C$, the relative change under compression cancels $C$ exactly, so the $+111\%$ pruning compounding claim holds whether a Fair-Speech caption error costs 2, 5, or 8 seconds; only the absolute anchors (12 to 102\,s/min across $C$) depend on the assumption. At five seconds per error, the worst-served Fair-Speech group already carries roughly 30 minutes of additional correction effort per hour of speech at full precision, and 50\% Wanda raises this to about 64 minutes: an extra half-hour of repair labor per hour of audio attributable to compression alone. This decoupling makes temporal taxation a useful operational fairness metric beyond mean WER \citep{choi2025temporal}: single-snapshot FP16 audits miss the redistribution that follows at deployment \citep{koenecke2020racial,veliche2024fairspeech,ferraz2024multilingual}, so deployment-time audits should evaluate at least one sub-8-bit precision and one pruned condition.

\section*{Limitations}
\label{sec:limitations}

The main results cover one model family, Whisper; a comprehensive cross-family audit is left to future work.

The evaluation is English-only and uses read-speech corpora exclusively (voice-assistant prompts, read sentences, and read clinical text). Post-training weight compression may compound differently on tonal or agglutinative languages. The read-speech caveat is asymmetric in a known direction: the WER gap between spontaneous and read speech is largest for AAVE-speaking and Black/AA populations \citep{martin2020habitual,cunningham2024impacts}, and these are also the groups carrying the largest compounding effect in our experiments. The $+111\%$ Wanda compounding result on Fair-Speech is therefore plausibly a lower bound on the corresponding effect in spontaneous AAVE; read-speech evaluation works against detection of the disparity, not toward it.

Audio-quality provenance remains a partial confound. The SNR control in Section~\ref{sec:pruning:snr} suggests that recording quality is not the source of the demographic gap, but SNR is a partial proxy for recording quality: it does not capture reverberation or non-linear distortion, and a perceptual-quality control (for example DNSMOS) was not run. A residual confound between group membership and recording conditions therefore cannot be ruled out without controlled recording-environment data \citep{li2024provenance}.

Each compression method is evaluated through a single reference implementation, so results are specific to that implementation rather than the underlying algorithm. GPTQ, AWQ, and SparseGPT are excluded because the reference libraries we tested expose per-architecture calibration classes that do not include encoder-decoder speech models, not because the algorithms are architecture-specific \citep{frantar2023gptq,lin2023awq,frantar2023sparsegpt}; Wanda 2:4 structured pruning was attempted and produced catastrophic outputs (WER 194\% on a 20-utterance test with 32 calibration utterances). Pruning is therefore evaluated only at unstructured 50\% sparsity: on all five Whisper backbones for Fair-Speech and Common Voice 25, and on \textit{whisper-tiny}, \textit{whisper-base}, and \textit{whisper-medium} for AfriSpeech. All compression is strictly post-training; quantization-aware training and demographic-aware retraining are natural extensions left to future work.

All results are measured on single released checkpoints. Our contrasts are within-checkpoint, comparing FP16 against compressed versions of identical weights, so variance attributable to the original training run largely differences out. What remains unmeasured is whether the magnitude of the compounding effect would replicate across independently trained checkpoints. Group fairness metrics are known to vary substantially across training runs in other modalities \citep{ganesh2023impact}, and we did not retrain any Whisper model.

Four additional experiments were proposed during review and are not included here: an AWQ arm configured through torchao, a DNSMOS perceptual-quality control on the \textit{whisper-large-v3} + Wanda + Fair-Speech cell, an FP8 (E4M3) arm paralleling INT8, and a search-over-scale variant of INT8 on \textit{whisper-large-v3} + Fair-Speech. None were completed in time for this version. The first would test whether an activation-aware 4-bit recipe redistributes error differently from the two data-free recipes evaluated here; the second would replace the SNR proxy with a perceptual measure; the third and fourth would probe how sensitive the INT8-is-gentle result is to number format and scale selection. We record their absence here rather than leaving the commitments unaddressed; the conclusions of Sections~\ref{sec:pruning} and~\ref{sec:quantization} should be read as covering NF4, HQQ, INT8 at library-default scale, and Wanda 50\% unstructured only.

\section*{Ethical considerations}
\label{sec:ethics}

All datasets are used under their published licenses, with no modification and no new data collection; demographic categories follow speaker self-identification in the source corpora. The compute footprint is approximately 150 A100 GPU-hours. Code, configurations, and result tables will be released on publication.

A foreseeable misreading is that compounding argues against deploying compressed ASR for marginalized populations. The intended reading is the opposite: withholding edge-deployable ASR would itself disadvantage the same speakers, since edge precision is often the only feasible delivery channel. These findings motivate fairness-aware compression and per-deployment audits, not a retreat from compression.

\bibliography{refs}

\begin{thebibliography}{30}
\providecommand{\natexlab}[1]{#1}

\bibitem[{Ahia et~al.(2021)Ahia, Kreutzer, and Hooker}]{ahia2021lowresource}
Orevaoghene Ahia, Julia Kreutzer, and Sara Hooker. 2021.
\newblock \href {https://arxiv.org/abs/2110.03036} {The low-resource double
  bind: An empirical study of pruning for low-resource machine translation}.
\newblock In \emph{Findings of the Association for Computational Linguistics:
  EMNLP 2021}.

\bibitem[{Andreyev(2025)}]{andreyev2025whisperquant}
Allison Andreyev. 2025.
\newblock \href {https://arxiv.org/abs/2503.09905} {Quantization for {OpenAI}'s
  whisper models: A comparative analysis}.
\newblock \emph{Preprint}, arXiv:2503.09905.

\bibitem[{Badri and Shaji(2023)}]{badri2023hqq}
Hicham Badri and Appu Shaji. 2023.
\newblock \href {https://mobiusml.github.io/hqq_blog/} {Half-quadratic
  quantization of large machine learning models}.
\newblock Mobius Labs blog post.

\bibitem[{Benjamini and Hochberg(1995)}]{benjamini1995controlling}
Yoav Benjamini and Yosef Hochberg. 1995.
\newblock \href {https://doi.org/10.1111/j.2517-6161.1995.tb02031.x}
  {Controlling the false discovery rate: A practical and powerful approach to
  multiple testing}.
\newblock \emph{Journal of the Royal Statistical Society: Series B
  (Methodological)}, 57(1):289-300.

\bibitem[{Cheng et~al.(2026)Cheng, Clemmensen, and Das}]{cheng2026diversity}
Ting-Hui Cheng, Line~H. Clemmensen, and Sneha Das. 2026.
\newblock \href {https://arxiv.org/abs/2603.05267} {Beyond word error rate:
  Auditing the diversity tax in speech recognition through dataset
  cartography}.
\newblock \emph{Preprint}, arXiv:2603.05267.

\bibitem[{Choi and Choi(2025)}]{choi2025temporal}
Anna Seo~Gyeong Choi and Hoon Choi. 2025.
\newblock \href {https://arxiv.org/abs/2508.07143} {Fairness of automatic
  speech recognition: Looking through a philosophical lens}.
\newblock In \emph{Proceedings of the AAAI/ACM Conference on AI, Ethics, and
  Society (AIES)}.

\bibitem[{Cunningham et~al.(2024)Cunningham, Blodgett, Madaio, Daum{\'e}~III,
  Harrington, and Wallach}]{cunningham2024impacts}
Jay Cunningham, Su~Lin Blodgett, Michael Madaio, Hal Daum{\'e}~III, Christina
  Harrington, and Hanna Wallach. 2024.
\newblock \href {https://doi.org/10.18653/v1/2024.findings-acl.761}
  {Understanding the impacts of language technologies' performance disparities
  on {A}frican {A}merican language speakers}.
\newblock In \emph{Findings of the Association for Computational Linguistics:
  ACL 2024}, pages 12826-12833, Bangkok, Thailand. Association for
  Computational Linguistics.

\bibitem[{Dettmers et~al.(2022)Dettmers, Lewis, Belkada, and
  Zettlemoyer}]{dettmers2022llmint8}
Tim Dettmers, Mike Lewis, Younes Belkada, and Luke Zettlemoyer. 2022.
\newblock \href {https://arxiv.org/abs/2208.07339} {{LLM.int8()}: 8-bit matrix
  multiplication for transformers at scale}.
\newblock In \emph{Advances in Neural Information Processing Systems
  (NeurIPS)}.

\bibitem[{Ezema et~al.(2025)Ezema, Chandler, Southwell, Cholendiran, and
  D'Mello}]{ezema2025cascading}
Kelechi Ezema, Chelsea Chandler, Rosy Southwell, Nimal Cholendiran, and Sidney
  D'Mello. 2025.
\newblock \href {https://doi.org/10.1145/3706598.3714059} {{``It Feels like
  We're Not Meeting the Criteria'': Examining and Mitigating the Cascading
  Effects of Bias in Automatic Speech Recognition in Spoken Language
  Interfaces}}.
\newblock In \emph{Proceedings of the 2025 CHI Conference on Human Factors in
  Computing Systems}, pages 1-13. Association for Computing Machinery.

\bibitem[{Feng et~al.(2025)Feng, Lin, Zhuo, Su, Ramakrishnan, Yuan, and
  Zhang}]{feng2025edgeasr}
Chen Feng, Yicheng Lin, Shaojie Zhuo, Chenzheng Su, Ramchalam~Kinattinkara
  Ramakrishnan, Zhaocong Yuan, and Xiaopeng Zhang. 2025.
\newblock \href {https://arxiv.org/abs/2507.07877} {{Edge-ASR}: Towards low-bit
  quantization of automatic speech recognition models}.
\newblock \emph{Preprint}, arXiv:2507.07877.

\bibitem[{Ferraz(2024)}]{ferraz2024multilingual}
Thomas~Palmeira Ferraz. 2024.
\newblock \href {https://arxiv.org/abs/2405.00966} {Efficient compression of
  multitask multilingual speech models}.
\newblock \emph{Preprint}, arXiv:2405.00966.

\bibitem[{Frantar and Alistarh(2023)}]{frantar2023sparsegpt}
Elias Frantar and Dan Alistarh. 2023.
\newblock \href {https://proceedings.mlr.press/v202/frantar23a.html}
  {{SparseGPT}: Massive language models can be accurately pruned in one-shot}.
\newblock In \emph{Proceedings of the 40th International Conference on Machine
  Learning}, volume 202 of \emph{Proceedings of Machine Learning Research},
  pages 10323-10337. PMLR.

\bibitem[{Frantar et~al.(2023)Frantar, Ashkboos, Hoefler, and
  Alistarh}]{frantar2023gptq}
Elias Frantar, Saleh Ashkboos, Torsten Hoefler, and Dan Alistarh. 2023.
\newblock \href {https://arxiv.org/abs/2210.17323} {{GPTQ}: Accurate
  post-training quantization for generative pre-trained transformers}.
\newblock In \emph{Proceedings of the Eleventh International Conference on
  Learning Representations (ICLR)}.

\bibitem[{Gandhi et~al.(2023)Gandhi, von Platen, and
  Rush}]{gandhi2023distilwhisper}
Sanchit Gandhi, Patrick von Platen, and Alexander~M. Rush. 2023.
\newblock \href {https://arxiv.org/abs/2311.00430} {Distil-whisper: Robust
  knowledge distillation via large-scale pseudo labelling}.
\newblock \emph{Preprint}, arXiv:2311.00430.

\bibitem[{Ganesh et~al.(2023)Ganesh, Chang, Strobel, and
  Shokri}]{ganesh2023impact}
Prakhar Ganesh, Hongyan Chang, Martin Strobel, and Reza Shokri. 2023.
\newblock \href {https://doi.org/10.1145/3593013.3594116} {On the impact of
  machine learning randomness on group fairness}.
\newblock In \emph{Proceedings of the 2023 ACM Conference on Fairness,
  Accountability, and Transparency}, pages 1789-1800. Association for
  Computing Machinery.

\bibitem[{Ginjala et~al.(2026)Ginjala, Fosler-Lussier, Myers, and
  Parthasarathy}]{ginjala2026prior}
Srishti Ginjala, Eric Fosler-Lussier, Christopher~W. Myers, and Srinivasan
  Parthasarathy. 2026.
\newblock \href {https://arxiv.org/abs/2604.21276} {Do {LLM} decoders listen
  fairly? benchmarking how language model priors shape bias in speech
  recognition}.
\newblock \emph{Preprint}, arXiv:2604.21276.

\bibitem[{Harris et~al.(2024)Harris, Mgbahurike, Kumar, and
  Yang}]{harris2024modeling}
Camille Harris, Chijioke Mgbahurike, Neha Kumar, and Diyi Yang. 2024.
\newblock \href {https://doi.org/10.18653/v1/2024.findings-emnlp.890} {Modeling
  gender and dialect bias in automatic speech recognition}.
\newblock In \emph{Findings of the Association for Computational Linguistics:
  EMNLP 2024}, pages 15166-15184, Miami, Florida, USA. Association for
  Computational Linguistics.

\bibitem[{Hooker et~al.(2019)Hooker, Courville, Clark, Dauphin, and
  Frome}]{hooker2019compressed}
Sara Hooker, Aaron Courville, Gregory Clark, Yann Dauphin, and Andrea Frome.
  2019.
\newblock \href {https://arxiv.org/abs/1911.05248} {What do compressed deep
  neural networks forget?}
\newblock \emph{Preprint}, arXiv:1911.05248.

\bibitem[{Hooker et~al.(2020)Hooker, Moorosi, Clark, Bengio, and
  Denton}]{hooker2020characterising}
Sara Hooker, Nyalleng Moorosi, Gregory Clark, Samy Bengio, and Emily Denton.
  2020.
\newblock \href {https://arxiv.org/abs/2010.03058} {Characterising bias in
  compressed models}.
\newblock \emph{Preprint}, arXiv:2010.03058.

\bibitem[{Kirsten et~al.(2025)Kirsten, Habernal, Nanda, and
  Zafar}]{kirsten2025inference}
Elisabeth Kirsten, Ivan Habernal, Vedant Nanda, and Muhammad~Bilal Zafar. 2025.
\newblock \href {https://arxiv.org/abs/2410.22118} {The impact of inference
  acceleration on bias of {LLM}s}.
\newblock In \emph{Proceedings of the 2025 Conference of the Nations of the
  Americas Chapter of the Association for Computational Linguistics: Human
  Language Technologies (Volume 1: Long Papers)}, pages 1834-1853,
  Albuquerque, New Mexico. Association for Computational Linguistics.

\bibitem[{Koenecke et~al.(2024)Koenecke, Choi, Mei, Schellmann, and
  Sloane}]{koenecke2024careless}
Allison Koenecke, Anna Seo~Gyeong Choi, Katelyn~X. Mei, Hilke Schellmann, and
  Mona Sloane. 2024.
\newblock \href {https://doi.org/10.1145/3630106.3658996} {Careless whisper:
  Speech-to-text hallucination harms}.
\newblock In \emph{Proceedings of the 2024 ACM Conference on Fairness,
  Accountability, and Transparency (FAccT)}.

\bibitem[{Koenecke et~al.(2020)Koenecke, Nam, Lake, Nudell, Quartey, Mengesha,
  Toups, Rickford, Jurafsky, and Goel}]{koenecke2020racial}
Allison Koenecke, Andrew Nam, Emily Lake, Joe Nudell, Minnie Quartey, Zion
  Mengesha, Connor Toups, John~R. Rickford, Dan Jurafsky, and Sharad Goel.
  2020.
\newblock \href {https://doi.org/10.1073/pnas.1915768117} {Racial disparities
  in automated speech recognition}.
\newblock \emph{Proceedings of the National Academy of Sciences},
  117(14):7684-7689.

\bibitem[{Li et~al.(2024)Li, Cohen, and Pakhomov}]{li2024provenance}
Changye Li, Trevor Cohen, and Serguei Pakhomov. 2024.
\newblock \href {https://arxiv.org/abs/2407.13982} {Reexamining racial
  disparities in automatic speech recognition performance: The role of
  confounding by provenance}.
\newblock \emph{Preprint}, arXiv:2407.13982.

\bibitem[{Lin et~al.(2024)Lin, Tang, Tang, Yang, Chen, Wang, Xiao, Dang, Gan,
  and Han}]{lin2023awq}
Ji~Lin, Jiaming Tang, Haotian Tang, Shang Yang, Wei-Ming Chen, Wei-Chen Wang,
  Guangxuan Xiao, Xingyu Dang, Chuang Gan, and Song Han. 2024.
\newblock \href {https://arxiv.org/abs/2306.00978} {{AWQ}: Activation-aware
  weight quantization for {LLM} compression and acceleration}.
\newblock In \emph{Proceedings of Machine Learning and Systems (MLSys)}.

\bibitem[{Martin and Tang(2020)}]{martin2020habitual}
Joshua~L. Martin and Kevin Tang. 2020.
\newblock \href {https://doi.org/10.21437/Interspeech.2020-2893} {Understanding
  racial disparities in automatic speech recognition: The case of habitual
  ``be''}.
\newblock In \emph{Proceedings of Interspeech 2020}, pages 626-630.

\bibitem[{Ogueji et~al.(2022)Ogueji, Ahia, Onilude, Gehrmann, Hooker, and
  Kreutzer}]{ogueji2022compression}
Kelechi Ogueji, Orevaoghene Ahia, Gbemileke Onilude, Sebastian Gehrmann, Sara
  Hooker, and Julia Kreutzer. 2022.
\newblock \href {https://arxiv.org/abs/2211.02738} {Intriguing properties of
  compression on multilingual models}.
\newblock In \emph{Proceedings of the 2022 Conference on Empirical Methods in
  Natural Language Processing (EMNLP)}, pages 9092-9110, Abu Dhabi, United
  Arab Emirates. Association for Computational Linguistics.

\bibitem[{Radford et~al.(2023)Radford, Kim, Xu, Brockman, McLeavey, and
  Sutskever}]{radford2023whisper}
Alec Radford, Jong~Wook Kim, Tao Xu, Greg Brockman, Christine McLeavey, and
  Ilya Sutskever. 2023.
\newblock \href {https://arxiv.org/abs/2212.04356} {Robust speech recognition
  via large-scale weak supervision}.
\newblock In \emph{Proceedings of the 40th International Conference on Machine
  Learning (ICML)}.

\bibitem[{Sun et~al.(2024)Sun, Liu, Bair, and Kolter}]{sun2024wanda}
Mingjie Sun, Zhuang Liu, Anna Bair, and J.~Zico Kolter. 2024.
\newblock \href {https://arxiv.org/abs/2306.11695} {A simple and effective
  pruning approach for large language models}.
\newblock In \emph{Proceedings of the Twelfth International Conference on
  Learning Representations (ICLR)}.

\bibitem[{Tatman and Kasten(2017)}]{tatman2017gender}
Rachael Tatman and Conner Kasten. 2017.
\newblock \href {https://doi.org/10.21437/Interspeech.2017-1746} {Effects of
  talker dialect, gender \& race on accuracy of {Bing Speech} and {YouTube}
  automatic captions}.
\newblock In \emph{Proceedings of Interspeech 2017}, pages 934-938.

\bibitem[{Veliche et~al.(2024)Veliche, Huang, Kochaniyan, Peng, Kalinli, and
  Seltzer}]{veliche2024fairspeech}
Irina-Elena Veliche, Zhuangqun Huang, Vineeth~Ayyat Kochaniyan, Fuchun Peng,
  Ozlem Kalinli, and Michael~L. Seltzer. 2024.
\newblock \href {https://arxiv.org/abs/2408.12734} {Towards measuring fairness
  in speech recognition: {Fair-Speech} dataset}.
\newblock \emph{Preprint}, arXiv:2408.12734.

\end{thebibliography}

\appendix

\section{Ranking stability and per-cell significance}
\label{app:full-results}
\label{app:kendall}

\begin{table}[h]
  \small
  \centering
  \begin{tabular}{lll}
    \toprule
    Dataset & Precision & Kendall $\tau$ \\
    \midrule
    AfriSpeech    & INT8     & 1.000 \\
    AfriSpeech    & INT4 NF4 & 1.000 \\
    AfriSpeech    & INT4 HQQ & 1.000 \\
    Common Voice 25 & INT8     & 1.000 \\
    Common Voice 25 & INT4 NF4 & 1.000 \\
    Common Voice 25 & INT4 HQQ & 1.000 \\
    Common Voice 25 & Wanda 50\% & 1.000 \\
    Fair-Speech   & INT8     & 1.000 \\
    Fair-Speech   & INT4 NF4 & 1.000 \\
    Fair-Speech   & INT4 HQQ & \textbf{0.929} \\
    Fair-Speech   & Wanda 50\% & 1.000 \\
    \bottomrule
  \end{tabular}
  \caption{Kendall's $\tau$ between FP16 model rankings by MMR and rankings under each compressed precision. The single adjacent-pair swap at Fair-Speech + INT4 HQQ is the only deviation from a perfect ordering.}
  \label{tab:kendall}
\end{table}

Aggregated BH-FDR significance counts ($q < 0.05$ within each cell, paired permutation vs.\ FP16) over the 56 Common Voice 25 (model, accent) pairs are: INT8, 1 of 56 significant per-accent deltas (\textit{distil-large-v3}, us, $+0.40$\,pp); INT4 NF4, 11 of 56 (concentrated on \textit{whisper-tiny} and \textit{whisper-base}); INT4 HQQ, 12 of 56 (the \textit{whisper-tiny} + Common Voice 25 cell carries all seven accent deltas at $q < 0.05$). Counting whole cells instead, at least one per-group delta reaches $q<0.05$ in 6 of 21 INT8 cells, 18 of 21 INT4 NF4 cells, 15 of 21 INT4 HQQ cells, and 13 of 13 Wanda 50\% cells. Across all three datasets INT8 produces 7 significant per-group deltas out of 427. Four of the seven fall on AfriSpeech accents with $n \leq 101$ (bini $+3.35$\,pp, khana $+3.05$\,pp, etche $+2.64$\,pp, mwaghavul $+1.80$\,pp); on the two larger benchmarks the significant deltas are at most $+0.40$\,pp (Common Voice 25 us) and $+0.22$\,pp (Fair-Speech Black/AA). INT8 is close to, but not exactly, fairness-neutral, and its exceptions sit on the smallest accent strata. The Wanda + \textit{whisper-large-v3} + Fair-Speech result is concentrated rather than uniform: Black/AA carries $+7.89$\,pp ($p_{\mathrm{BH}} \approx 0$) while the next-largest delta is more than five times smaller, Middle Eastern at $+1.50$\,pp, followed by Pacific Islander at $+1.47$\,pp; Hispanic ($+0.38$\,pp, $p_{\mathrm{BH}} = 0.15$) and White ($-0.03$\,pp) fall below the BH threshold. Decomposing the same cell, deletion rate accounts for 61\% of the Black/AA WER increase and only 34\% of the Asian increase, consistent with a long-tail account of how Wanda's magnitude $\times$ activation criterion preferentially removes weights serving rarer input distributions \citep{hooker2019compressed}.

\section{Temporal-taxation sensitivity}
\label{app:taxation-sensitivity}

Because $T_g = \mathrm{WER}_g \cdot C \cdot \mathit{wpm}$ is linear in $C$, the worst-vs-best differential scales linearly with $C$ and the relative change under compression cancels $C$ exactly. Table~\ref{tab:tax-sens} reports the top ten cells by relative compounding at $C \in \{2, 5, 8\}$\,s/edit. Relative change is identical at every $C$; absolute anchors scale by $C/5$ from the middle column. Nine of the top ten cells are Wanda; the only sub-8-bit quantization cell to enter the top ten is \textit{whisper-tiny} + INT4 HQQ on CV25.

\begin{table*}[t]
  \small
  \centering
  \begin{tabular}{llrrrr}
    \toprule
    Model & Cell & $C{=}2$ & $C{=}5$ & $C{=}8$ & rel.\\
    \midrule
    whisper-large-v3 & Wanda, FS Black/AA vs Asian       & 12.06$\to$25.49 & 30.14$\to$63.73 & 48.23$\to$101.97 & $+111\%$ \\
    whisper-medium   & Wanda, FS Black/AA vs Asian       & 14.99$\to$28.21 & 37.46$\to$70.53 & 59.94$\to$112.85 & $+88\%$ \\
    whisper-base     & Wanda, CV25 indian vs canada      & 29.86$\to$50.79 & 74.64$\to$126.97 & 119.43$\to$203.16 & $+70\%$ \\
    whisper-large-v3 & Wanda, CV25 indian vs canada      & 14.08$\to$23.79 & 35.20$\to$59.47 & 56.32$\to$95.15  & $+69\%$ \\
    whisper-small    & Wanda, FS Black/AA vs Asian       & 23.48$\to$35.67 & 58.69$\to$89.17 & 93.91$\to$142.68 & $+52\%$ \\
    whisper-tiny     & INT4 HQQ, CV25 african vs canada  & 38.82$\to$53.99 & 97.04$\to$134.97 & 155.27$\to$215.95 & $+39\%$ \\
    whisper-small    & Wanda, CV25 african vs canada     & 27.64$\to$37.33 & 69.10$\to$93.33 & 110.56$\to$149.32 & $+35\%$ \\
    whisper-medium   & Wanda, CV25 indian vs canada      & 18.48$\to$24.41 & 46.19$\to$61.03 & 73.91$\to$97.66  & $+32\%$ \\
    whisper-tiny     & Wanda, CV25 african vs canada     & 38.82$\to$50.10 & 97.04$\to$125.26 & 155.27$\to$200.41 & $+29\%$ \\
    whisper-base     & Wanda, FS Black/AA vs Asian       & 40.12$\to$49.63 & 100.30$\to$124.09 & 160.48$\to$198.54 & $+24\%$ \\
    \bottomrule
  \end{tabular}
  \caption{Worst-vs-best temporal-taxation differential in s/min, FP16 $\to$ compressed, at three cost-per-edit values. The relative change column is invariant in $C$.}
  \label{tab:tax-sens}
\end{table*}

\section{SNR confound check}
\label{app:snr}

Following \citet{li2024provenance}, we compute a per-utterance SNR proxy on every Fair-Speech utterance, regress per-utterance WER on SNR via ordinary least squares, and re-test the demographic F-statistic on the residual. Every one of the 34 (model, precision) Fair-Speech cells remains demographically significant after the SNR control ($F$-test $p < 0.05$). The MMR after SNR adjustment is larger than the raw MMR in 34 of 34 cells, on average 15.0\% larger (median 13.9\%; range 4.7\% to 32.3\% larger across cells). The SNR control therefore widens the gap rather than narrowing it, so the raw corpus WER mildly understates the disparity.

\section{Loop-flag threshold and reproducibility}
\label{app:determinism}

The per-utterance loop flag triggers when a hypothesis contains more than five 5-gram repetitions or its length exceeds three times the reference length. Both thresholds were fixed before the full sweep was run, based on a preliminary inspection of 100 utterances in which non-pathological hypotheses had at most five 5-gram repetitions (intentional repetition in voice-assistant prompts) and a length ratio of at most 2.7.

\paragraph{Threshold sensitivity.} To verify that the loop-rate compounding result does not hinge on these specific cutoffs, we re-derive the loop flag at two perturbations of the (rep5, length-ratio) pair: a loose setting $(4, 2.5)$ and a strict setting $(6, 3.5)$. The recomputation is run directly on the stored per-utterance \texttt{rep5} and \texttt{len\_ratio} columns and requires no re-inference. Table~\ref{tab:loop-sens} reports the H2 headline cell (\textit{whisper-tiny}~+~INT4 HQQ on AfriSpeech) at each setting. Aggregate loop rate, BH-significant accent count, and the compounding-index sign are stable. Three of the five headline West African accents (Kanuri, Yoruba, Hausa) clear BH-FDR significance at every threshold (Table~\ref{tab:loop-sens-pg}). Swahili reaches significance only at the strict setting; Igbo at none. Per-accent loop-rate ratios for these three accents stay within a 3.5 to 6.7-fold band across the threshold range.

\begin{table}[t]
  \small
  \centering
  \begin{tabular}{lrrr}
    \toprule
    (rep5, len-ratio) & FP16$\to$HQQ & BH-sig & CI \\
    \midrule
    Loose $(4, 2.5)$    & 2.07\% $\to$ 6.79\% & 6/62 & $+0.73$ \\
    Baseline $(5, 3.0)$ & 1.50\% $\to$ 5.78\% & 7/62 & $+0.68$ \\
    Strict $(6, 3.5)$   & 1.42\% $\to$ 5.46\% & 8/62 & $+0.56$ \\
    \bottomrule
  \end{tabular}
  \caption{H2 cell (\textit{whisper-tiny}~+~INT4 HQQ on AfriSpeech) under three loop-flag thresholds. BH-sig is the count of accents (out of 62 with $n \geq 30$) reaching $q < 0.05$ in two-proportion z-tests with Benjamini-Hochberg correction. CI is the smoothed loop-rate MMR compounding index. Aggregate loop rate decreases as the threshold tightens, but the FP16-to-HQQ contrast and the per-accent ranking are preserved.}
  \label{tab:loop-sens}
\end{table}

\begin{table}[t]
  \small
  \centering
  \begin{tabular}{lrrr}
    \toprule
    Accent ($n$) & Loose & Baseline & Strict \\
    \midrule
    Kanuri (66)   & $4.00\times^{\star}$ & $6.67\times^{\star}$ & $6.33\times^{\star}$ \\
    Yoruba (648)  & $3.50\times^{\star}$ & $5.12\times^{\star}$ & $5.00\times^{\star}$ \\
    Hausa (196)   & $4.75\times^{\star}$ & $6.00\times^{\star}$ & $5.00\times^{\star}$ \\
    Swahili (521) & $2.71\times$         & $3.40\times$         & $4.25\times^{\star}$ \\
    Igbo (355)    & $2.20\times$         & $2.83\times$         & $2.83\times$ \\
    \bottomrule
  \end{tabular}
  \caption{INT4 HQQ over FP16 loop-rate ratio under \textit{whisper-tiny} on AfriSpeech, for the five headline West African accents at each loop-flag setting. $\star$ marks $q < 0.05$ after BH-FDR correction within the 62-accent slate. The three accents driving the H2 claim (Kanuri, Yoruba, Hausa) are significant at every setting.}
  \label{tab:loop-sens-pg}
\end{table}

All inference uses greedy decoding (except for the beam-search ablation in Section~\ref{sec:beam}) with deterministic-kernel flags, fixed random seeds, and a pinned environment. Two reruns of the same 100 utterance ids on Whisper-large-v3 + FP16 produced bit-identical hypotheses on the same hardware.

\section{Excluded compression methods}
\label{app:excluded}

AWQ \citep{lin2023awq} is excluded because its reference library has been deprecated and its successor pipeline is tied to a serving runtime without first-class encoder-decoder support. GPTQ \citep{frantar2023gptq} and SparseGPT \citep{frantar2023sparsegpt} are excluded because their reference implementations expose per-architecture classes that do not currently include any encoder-decoder speech model; implementing a Whisper adapter for either method would require non-trivial new infrastructure (per-block calibration walks across encoder, decoder, and cross-attention) and would not add a contrast beyond the data-free 4-bit recipes (NF4 and HQQ) already evaluated. Wanda 2:4 structured pruning was attempted on \textit{whisper-small} and produced catastrophic outputs on a 20-utterance test with 32 calibration utterances (WER 194\%), incompatible with the narrow attention-head dimensions of Whisper, and is therefore dropped from the main sweep.

In every case the blocker is the reference implementation rather than the algorithm. The specific blockers we hit were: \texttt{gptqmodel} 7.0 lists no encoder-decoder architecture among its supported model classes and requires a \texttt{transformers} major-version upgrade that would invalidate our pinned determinism baseline; \texttt{auto-gptq} does not build against the pinned torch and Python versions; and \texttt{autoawq} fails to import against \texttt{transformers} 4.57 and has been deprecated in favor of a serving-runtime pipeline. A Whisper adapter for any of these would need per-block calibration walks across the encoder, the decoder, and the decoder cross-attention projections. We note for future work that torchao exposes a module-level configuration path for AWQ that does not require a per-architecture class, which we did not pursue for this version.

\section{Granite-4.0 generalization probe}
\label{app:granite}

To check whether the INT8 portion of the quantization findings generalizes beyond Whisper, we evaluate IBM Granite-4.0-1b-speech (a 1B-parameter speech model with architecture distinct from Whisper) on Fair-Speech at FP16 and INT8. The probe is appendix-only and cannot test the main claims of this paper: pruning was not applied to Granite, and INT4 was not evaluated, so the probe speaks only to the INT8 sub-claim of Section~\ref{sec:quantization} and to the baseline demographic-gap structure.

Granite-Speech is less accurate than Whisper-large-v3 at FP16 on Fair-Speech (utterance-weighted mean WER including loop-flagged hypotheses, 10.72\% vs 6.10\%), but the per-ethnicity gap structure is essentially identical (per-group figures below, as everywhere in the paper, are loop-filtered corpus WER): Black/AA speakers carry the highest WER (14.79\% on Granite, 9.78\% on Whisper-large-v3), Asian speakers the lowest (5.13\% on Granite, 3.27\% on Whisper-large-v3), and the FP16 max-min ratios across ethnicity groups are within $0.1$ of each other (Granite MMR $= 2.88$; Whisper-large-v3 MMR $= 2.99$). The shape of the demographic disparity therefore survives a substantial architectural change, which suggests the baseline gap is not a Whisper-specific training artifact and is more likely driven by training-data distribution shared across speech models or by Fair-Speech's own demographic composition.

INT8 quantization preserves both the aggregate accuracy and the demographic-gap structure on Granite. Aggregate WER moves from 10.72\% to 11.02\%, the MMR shifts from 2.883 to 2.896 (compounding index $+0.004$), loop rate is unchanged at 0.45\%, and the largest per-ethnicity WER delta is $+0.51$\,pp on Middle Eastern speakers ($n=747$). This is consistent with the Whisper INT8 result reported in Section~\ref{sec:quantization}, where INT8 produced 7 BH-significant per-group deltas out of 427 across all three datasets, none of them on Fair-Speech ethnicity groups larger than $+0.22$\,pp (Appendix~\ref{app:full-results}). The INT8-is-gentle finding therefore generalizes to one non-Whisper architecture. The probe does not adjudicate whether the pruning or INT4 findings generalize; cross-family experiments on those axes are left to future work.

\end{document}